# Medical Knowledge Embedding Based on Recursive Neural Network for Multi-Disease Diagnosis

Jingchi Jiang[1], Huanzheng Wang[1], Jing Xie[1], Xitong Guo[2], Yi Guan[1,*], and Qiubin Yu[3]

1. School of Computer Science and Technology, Harbin Institute of Technology, Integrated Laboratory Building 803, Harbin 150001, China. E-mail: jiangjingchi0118@163.com, whz123_hit@163.com, 16B903018@stu.hit.edu.cn, guanyi@hit.edu.cn.
2. School of Management, Harbin Institute of Technology, Harbin 150001, China. E-mail: xitongguo@hit.edu.cn.
3. Medical Record Room, Second Affiliated Hospital of Harbin Medical University, Harbin 150086, China. E-mail: yuqiubin6695@163.com.

\* Corresponding author

**Abstract**—The representation of knowledge based on first-order logic captures the richness of natural language and supports multiple probabilistic inference models. Although symbolic representation enables quantitative reasoning with statistical probability, it is difficult to utilize with machine learning models as they perform numerical operations. In contrast, knowledge embedding (*i.e.*, high-dimensional and continuous vectors) is a feasible approach to complex reasoning that can not only retain the semantic information of knowledge but also establish the quantifiable relationship among them. In this paper, we propose recursive neural knowledge network (RNKN), which combines medical knowledge based on first-order logic with recursive neural network for multi-disease diagnosis. After RNKN is efficiently trained from manually annotated Chinese Electronic Medical Records (CEMRs), diagnosis-oriented knowledge embeddings and weight matrixes are learned. Experimental results verify that the diagnostic accuracy of RNKN is superior to that of some classical machine learning models and Markov logic network (MLN). The results also demonstrate that the more explicit the evidence extracted from CEMRs is, the better is the performance achieved. RNKN gradually exhibits the interpretation of knowledge embeddings as the number of training epochs increases.

─ ─ ─ ─ ─ ─ ─ ─ ─ ◆ ─ ─ ─ ─ ─ ─ ─ ─ ─

## 1 INTRODUCTION

In clinical practice, medical knowledge plays a critical role and strongly reflects the diagnostical capability of clinicians. In recent years, there has been a rapid growth in ontologies and knowledge bases (KBs), such as Freebase [1], WordNet [2], and the Google Knowledge Graph [3], which store facts about real-world entities in the form of RDF triples. Similarly, Unified Medical Language System (UMLS) [4] knowledge sources collect metathesaurus, semantic network, and specialist lexicon to facilitate the development of intelligent systems in biomedicine and healthcare. As these KBs contain huge amounts of information, the question of how to represent the knowledge interpretably and participate in inferential tasks has become an increasingly important research issue.

Existing knowledge representation methods transform natural language into symbolic models, such as first-order formulas or triples of the form (*head entity*, *relation*, *tail entity*). Probabilistic graphical models (PGMs) [5-8] and knowledge graphs [9], which support uncertainty reasoning based on statistical relational learning, are concerned with the application of symbolic knowledge. Unfortunately, symbolic approaches have often suffered from the problem of entity linking [10-11] and word sense disambiguation [12-13] in practice. Moreover, the symbolic knowledge cannot be directly utilized in the numerical operation of machine learning.

Some knowledge representation approaches [14-19] address the concerns above by learning embedding that simulates the behavior of symbolic knowledge. For example, deep learning models such as word embedding [20-21], recurrent neural networks [22], or TransE [23] attempt to embed the relational knowledge into continuous vector spaces. Through training of a specified language model, the entity and relation embeddings that capture full syntactic and semantic richness of linguistic phrases are indirectly generated. Starting with Hinton and Williams [24], this idea has since been applied to statistical language modeling with considerable success. Mikolov et al. introduced the Skip-gram model [25], an efficient method for learning high-quality vector representations of words and phrases from large amounts of unstructured text data. Socher et al. proposes a neural tensor network (NTN) framework [26-27] for knowledge base completion in which additional facts are predicted from an existing database. Subsequent works [28-31] involved application to sentiment analysis, machine translation, and a wide range of NLP tasks. Learning interpretation representations of medical knowledge based on an ontological structure, graph-based attention model (GRAM) [32] exhibits promising performance for predictive modeling in healthcare.

However, the core argument against embeddings is inability to capture more complex patterns of reasoning such as those enabled by first-order logic [33]. In this study, to generate credible medical knowledge based on first-order logic, we identified medical entities and extracted entity relations from semi-structured Chinese electronic medical records

(CEMRs) [34]. Subsequently, the knowledge set of each CEMR was used to construct a Huffman tree. Further, we developed recursive neural knowledge network (RNKN), which combines medical knowledge based on first-order logic with recursive neural network for multi-disease diagnosis. In multi-disease diagnosis, RNKN not only preserves the reasoning ability of RNN, but also has expressive ability for logic knowledge. To evaluate the quality of the trained knowledge embeddings, we ascertained that the interpretable representations align well in lower-dimensional space.

The remainder of this paper is organized as follows. In Section 2, we introduce Chinese Electronic Medical Records and review the fundamentals of RNN and its optimization algorithms. In Section 3, the RNKN model is presented and its methodology described. In Section 4, we evaluate the effectiveness of RNKN for multi-disease diagnosis and further analyze the interpretation of medical knowledge representation. Finally, in Section 5, we conclude this paper and discuss directions for future work.

## 2 RELATED WORK

### 2.1 Chinese electronic medical records (CEMRs)

While some researchers have been focusing on building commonsense knowledge bases, others have been paying special attention to empirical knowledge, especially in the field of healthcare. In practice, clinicians are usually willing to accumulate professional knowledge and experience and apply the acquired knowledge in daily medical work. Discovery and application of empirical medical knowledge are equally important to intelligence diagnosis system. Since the reform of China's health system, CEMRs have gradually been formulated, making it possible to extract rich empirical medical knowledge. Therefore, we adopted these CEMRs from the Second Affiliated Hospital of Harbin Medical University as the primary source of medical knowledge. The types of CEMRs include discharge summary, progress note, doctor-patient agreement, and ultrasound report. In order to extract knowledge from them, we summarized the annotation guidelines of medical named entities and entity relationships in the CEMRs [35]. The entities are classified into four categories [36]: diseases, symptoms, tests, and treatments. In addition, we annotated seven assertions for the symptom entities and disease entities—specifically, present, absent, not associated with the patient, conditional, possible, historical, and occasional. The relationships are classified into five types and twelve classes, as shown in Table 1.

TABLE 1
TYPES AND CLASSES OF THE RELATIONSHIPS FOR SYMPTOM AND DISEASE ENTITIES

| Part A: Relation between "Disease" and "Symptom" | |
|---|---|
| Relation type | Example |
| Diseases cause symptoms (DCS) | 既往*脑梗死病史*10年, 遗留**左侧肢体活动受限** (*Cerebral infarction* at 10 years of age, remaining **activity limitation in the left extremities**) |
| Symptoms indicate diseases (SID) | **心悸**, **胸闷**, 伴**头晕**, 初步诊断为*冠心病* (With **palpitations**, **chest tightness**, **dizziness**, the preliminary diagnosis is *cardiovascular*) |

| Part B: Relation between "Treatment" and "Disease" | |
|---|---|
| Relation type | Example |
| Treatments improve diseases (TrID) | **高血压病**口服*利血平*控制, 可达到130-140/90 mmHg (Oral *reserpine* controls **hypertension**, and blood pressure can reach 130–140/90 mmHg) |
| Treatments worsen diseases (TrWD) | **高血压病**口服*替米沙坦*控制, 血压控制不佳 (Oral *telmisartan* controls **hypertension**, blood pressure was poorly controlled) |
| Treatments cause diseases (TrCD) | *电除颤*后, 引起**III度房室传导阻滞** (After *defibrillation*, **third degree atrioventricular block** resulted) |
| Treatments act on diseases (TrAD) | 既往**糖尿病史**20余年, *皮下注射胰岛素*控制 (**Type 2 diabetes** at 20 years of age was controlled by *subcutaneous injections of insulin*) |

| Part C: Relation between "Treatment" and "Symptom" | |
|---|---|
| Relation type | Example |
| Treatments improve symptoms (TrID) | 服用*钙剂*后, **后背部疼痛**显著缓解 (After taking *calcium*, **back pain** was significantly relieved) |
| Treatments worsen symptoms (TrWD) | 静点*青霉素*后, **咳嗽**加重 (After injection of *penicillin*, **cough** became even worse) |
| Treatments cause symptoms (TrCD) | 应用*派罗欣*后, 出现**体力下降**, **周身不适** (After applying *pegasys*, **physical strength decreased** and **general malaise** was presented) |
| Treatments act on symptoms (TrAD) | 无诱因出现**发热**, 口服*尼美舒利* (**Fever** appeared without cause, oral *nimesulide*) |

**Part D: Relation between "Test" and "Disease"**

| Relation type | Example |
|---|---|
| Tests confirm diseases (TeCD) | *心电图*: **心律失常**, **室性早搏** (*Electrocardiograms*: **arrhythmia**, **premature beat**) |

**Part E: Relation between "Test" and "Symptom"**

| Relation type | Example |
|---|---|
| Tests are adopted because of symptoms (TeAS) | 无诱因出现**发热**, 建议查*血常规* (**Fever** appeared without cause, *blood test* was recommended) |

## 2.2 Recursive neural network (RNN)

As a classical neural network framework, the standard RNN is applied to solve inductive inference tasks on complex symbolic structures of arbitrary size (such as logical terms, trees, or graphs). Fig. 1 illustrates this approach.

When a phrase is given, the RNN parses it into a binary semantic tree and computes the vector representation of each word. During the forward-propagation training period, the RNN computes parent vectors in a bottom-up fashion. The composition equation is as follows:

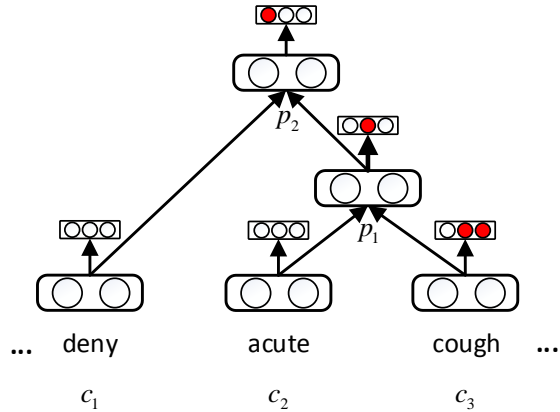

Fig. 1. Semantic tree of RNN models for classification task.

$$p_1 = f(W \begin{bmatrix} c_2 \\ c_3 \end{bmatrix} + b), \quad p_2 = f(W \begin{bmatrix} c_1 \\ p_1 \end{bmatrix} + b). \tag{1}$$

where $f$ is the activation function; $W \in \mathbb{R}^{d \times 2d}$ is the weight matrix, where $d$ is the dimensionality of the vector; and $b$ is the bias. Then, each parent vector $p_i$ is given as a feature to a softmax classifier such as that defined in Eq. 2 to compute its label probabilities:

$$y^p = softmax(W_s \cdot p) \tag{2}$$

where $W_s \in \mathbb{R}^{3 \times d}$ is the classification matrix. In this recursive process, the vector and classifying result of the node will gradually converge. After the vector of the leaf node is given, the RNN can ultimately map the semantic representation of the entire tree into the root vector.

In the above tree example, the classifying result of each node is represented as a disease vector $d = \{d_1, d_2, d_3\}$. When "cough" is manifested, $d_2$ and $d_3$ are considered as the possible diseases. By adding the modifier "acute," $d_2$ can be further confirmed. Until the negation word "deny" occurs, the output layer of the RNN presents the final classification result $d_1$, while a vector representation of the binary semantic tree is calculated as $p_2$.

Several improved RNN models [37-38] have been proposed. Of these, recursive neural tensor network (RNTN) [39] is the most typical. To reflect the significant interactions between the input vectors, RNTN introduces a tensor product into the composition function, which defines multiple bilinear forms. Its main contribution is that the binary tree not only contains richer semantic information, but also becomes more sensitive to error propagation.

## 3 METHODOLOGY

In this section, we present a general neural network framework for multi-relational representation learning. Although RNN and other neural network algorithms can be used for classification and word vectorization, the existing methods neglect the importance of domain knowledge. In the healthcare field, medical knowledge guides practitioners to diagnose the diseases and determine which treatment program is the best for which patient. In this context, we propose RNKN, which combines medical knowledge based on first-order logic with recursive neural network, as the answer to two concerning issues: "how can symbolic knowledge be represented in a neural network?" and "how can a neural network model be used for knowledge-based diagnosis?"

### 3.1 Recursive neural knowledge network

Suppose we are given a CEMR consisting of $n$ entities and $m$ relations. A set of triples comprising medical knowledge is constructed, which is stored as $\kappa = \{(e_i, r_k, e_j)\}$. The triple $(e_i, r_k, e_j)$ states that entity $e_i$ and entity $e_j$ are connected by relation $r_k$; e.g., the relation *SymptomIndicateDisease* can establish the relationship between symptom entity *Flustered* and disease entity *Angina*. The first-order logic form is *Flustered* ⇒ *Angina*. As illustrated in Fig. 2, our approach consists of four key components: 1) *We employ symptom and disease embedding as an input vector.* 2) *Medical knowledge base determines the connection between the input layer and the shallow logic layer $p^{(1)}$.* 3) *A Huffman tree based on the frequency of knowledge, which was inspired by word embedding models, is constructed to represent the deep logic knowledge.* 4) *Similar to RNN, we also use each node vector as features for a softmax classifier.*

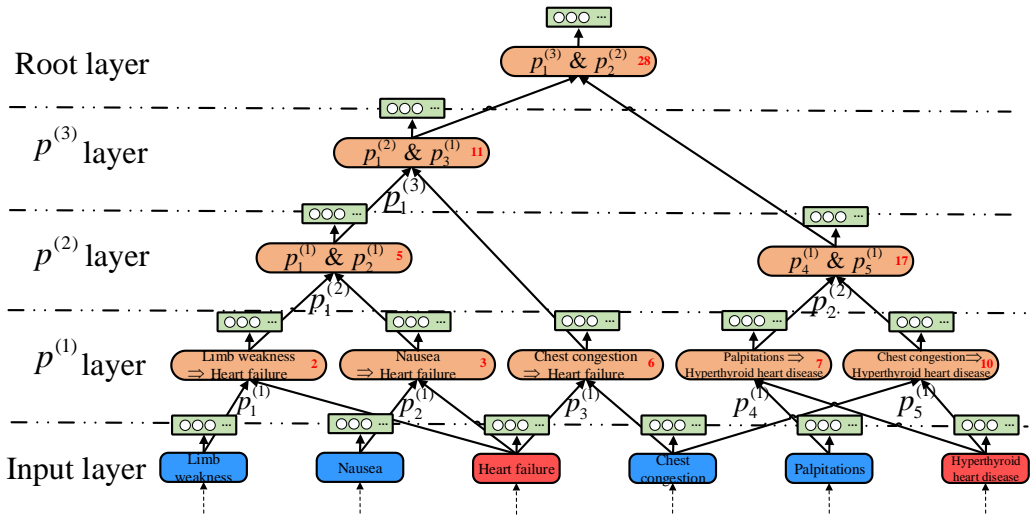

Fig. 2. A Huffman tree of RNKN represents deep first-order logic knowledge for multi-disease diagnosis.

Intuitively, a variety of logic symbols (e.g., &, ||, and ⇒) can also be expressed when we define the different weight matrix $W$. Meanwhile, by adjusting the hierarchical structures of the logic node of the Huffman tree, we can customize the logic operation to construct arbitrary first-order logic knowledge. Compared with the standard RNN, the main advantage is significant improvement of the interpretation of the hidden layers. Furthermore, the deep logic knowledge is innovatively used with the recursive neural network. The forward-propagation algorithm of RNKN is shown in detailed in Table 2.

TABLE 2
FORWARD-PROPAGATION ALGORITHM OF RNKN

| |
|---|
| **Algorithm 1**: Forward-propagation algorithm of RNKN |
| **Input:** $List_{EMR}$: list of electronic medical records for training. |
| **Output:** $List_{Vector}$: list of entity embeddings |
|     $Weight$: weight matrix |
|     $Weight_s$: disease classification matrix. |
| ***Begin*** |
| 1: **Initialize** the width of vector $d$ and the number of iterations $n$. |
| 2: **Extract** the entities and relationships from EMRs, $List_{EMR} \rightarrow List_{Knowledge}$. $List_{Knowledge} = \{<k_1, f_1>, <k_2, f_2>, ..., <k_n, f_n>\}$, where $k$ is the triple knowledge and $f$ is the frequency. |
| 3: **Initialize** the entity embeddings $List_{Vector}$, the weight matrix $Weight$, and the disease classification matrix $Weight_s$ in a random manner. |
| 4: ***for each*** $EMR_i \in List_{EMR}$ |
| 5:   Build a Huffman tree based on knowledge for each $EMR_i$, $EMR_i \rightarrow Tree_i$. |
| 6: ***end for*** |
| 7: ***while*** $loop < n$ |
| 8:   ***for*** $Tree_i \in List_{Tree}$ **do** |
| 9:     ForwardPropagation($Tree_i$.getRoot()) // Forward-propagation function of RNKN |
| 10:   ***end for*** |
| 11: ***end while*** |
| 12: ***Function*** ForwardPropagation($node$): |
| 13:   $child_{left} = node$.getLeftChildren() |
| 14:   $child_{right} = node$.getRightChildren() |
| 15:   ForwardPropagation ($child_{left}$) and ForwardPropagation ($child_{right}$) |
| 16:   **if** $child_{left}! = null\ \&\&\ child_{right}! = null$ **then** |
| 17:     $Vector_{combine}[v_1, ..., v_d] \leftarrow child_{left}$. getVector() |
| 18:     $Vector_{combine}[v_{d+1}, ..., v_{2d}] \leftarrow child_{right}$.getVector() |
| 19:     $Vector_{node} = f(Weight * Vector_{combine})$ |
| 20:   ***end if*** |
| 21: ***end Function*** |

## 3.2 Learning

In this section, we describe the back-propagation mechanism and the formulation of gradient descent. As shown in Fig. 2, each node has a softmax classifier trained on its vector representation to predict a target vector. In multi-disease diagnosis, the target vector is the probability distribution of the whole diagnosable disease $t \in \mathbb{R}^{C \times 1}$. Each element of the target vector is a normalization of co-occurrence frequency that the diagnosable disease and the logic knowledge are presented in the same CEMR. In order to minimize the cross-entropy error between the predicted distribution $y^i \in \mathbb{R}^{C \times 1}$ at node $i$ and its target distribution $t^i \in \mathbb{R}^{C \times 1}$, the error function of the RNKN for one CEMR is as follows:

$$E(\theta) = -\sum_i \sum_j t_j^i \log y_j^i + \lambda \|\theta\|^2 \quad (3)$$

Based on the definition of error vector $\delta_p \stackrel{\text{def}}{=} \partial E / \partial net_p$, the softmax error vector $\delta^{l,s}$ in the $p^{(l)}$ layer can be expressed as follows:

$$\delta^{l,s} = \delta^{l+1,s} \cdot W_s \cdot f'(x^l) \quad (4)$$

where $W_s$ is the softmax weight matrix, $f'$ is the element-wise derivative of activation function $f = \tanh$, and $x^l$ is the vector of node in layer $l$. To further calculate $\delta^{l+1,s}$ using chain rules, we transform it into the total derivative formulas.

$$\delta_i^{l+1,s} = \frac{\partial E}{\partial z_i^{l+1}} = \sum_{j=1}^n \frac{\partial E}{\partial y_j^{l+1}} \cdot \frac{\partial y_j^{l+1}}{\partial z_i^{l+1}} \quad (5)$$

From (3), we have $\partial E / \partial y_j = -t_j / y_j$. Knowing the expression of softmax as $y_j = e^{z_j} / \sum_{i=1}^n e^{z_i}$, the derivative of the predicted vector $y_j$ with respect to the weighted input vector $z_i$ is $\partial y_j / \partial z_i = y_i(1\{i = j\} - y_j)$. By substituting $\partial E / \partial y_j$ and $\partial y_j / \partial z_i$ into (5), the mathematical derivation is expressed as follows:

$$\delta_i^{l+1,s} = \frac{\partial E}{\partial z_i^{l+1}} = \sum_{j=1}^{n} \frac{\partial E}{\partial y_j^{l+1}} \cdot \frac{\partial y_j^{l+1}}{\partial z_i^{l+1}} = \sum_{j=1}^{n} (-\frac{t_j}{y_j}) \cdot y_i (1\{i=j\} - y_j)$$

$$= \left[ \sum_{\substack{j=1 \\ i \neq j}}^{n} (-\frac{t_j}{y_j}) \cdot y_i \cdot (-y_j) \right] + \left[ (-\frac{t_j}{y_j}) \cdot y_i \cdot (1 - y_i) \right] \quad . \quad (6)$$

$$= y_i \cdot \left[ \sum_{j=1}^{n} t_j - \frac{t_j}{y_j} \right] = y_i \cdot \left[ 1 - \frac{t_j}{y_j} \right]$$

$$= y_i - t_i$$

The rewritten expression of the softmax error vector is given as $\delta^{l,s} = (y_i - t_i) \cdot W_s \cdot f'(x^l)$. Similar to other RNN models, the error of RNKN consists of two parts: the softmax error shown above, and the error from the parent node into the leaf nodes through the tree structure, $\delta^{l,down}$. The sum of the two errors can be considered as the complete incoming error messages for node $i$ as $\delta_i^{l,com}$. Referring to the definition of $\delta^{l,down}$ in RNTN, we have the error message of $p_1^{(2)}$:

$$\delta^{p_1^{(2)},down} = \delta^{p_1^{(2)},com} \cdot W \cdot f'\left( \begin{bmatrix} p_1^{(1)} \\ p_2^{(1)} \end{bmatrix} \right). \quad (7)$$

Because $\delta^{p_1^{(2)},down} \in \mathbb{R}^{2d}$ is composed of two child nodes, we add half of $\delta^{p_1^{(2)},down}$ to the softmax error message of the corresponding child node. Consequently, we obtain the piecewise function of the complete error:

$$\delta^{i,com} = \begin{cases} \delta^{i,s} + \delta^{i,down}[1:d] & i \in leftchild \\ \delta^{i,s} + \delta^{i,down}[d+1:2d] & i \in rightchild \\ \delta^{i,s} & i \in root \end{cases} \quad (8)$$

Given the softmax error $\delta^{i,s}$, the structural error $\delta^{i,down}$, and the complete error $\delta^{i,com}$, we can deduce in detail the derivative of error function $E$ with respect to weight matrix $W$, softmax weight matrix $W_s$, and input layer vector $x$, respectively. In particular, we demonstrate the gradients at node $p_1^{(2)}$:

$$\begin{cases} \frac{\partial E}{\partial W_{p_1^{(2)}}} = \frac{\partial E}{\partial z_{p_1^{(2)}}} \frac{\partial z_{p_1^{(2)}}}{\partial W_{p_1^{(2)}}} = \delta^{p_1^{(2)},com} \cdot \begin{bmatrix} p_1^{(1)} \\ p_2^{(1)} \end{bmatrix} \\ \frac{\partial E}{\partial W_{s,p_1^{(2)}}} = \frac{\partial E}{\partial z_{p_1^{(2)}}^{l+1}} \frac{\partial z_{p_1^{(2)}}^{l+1}}{\partial W_{s,p_1^{(2)}}} = (y_i - t_i) \cdot f(x_{p_1^{(2)}}) \\ \frac{\partial E}{\partial x_{p_1^{(2)}}} = \delta^{p_1^{(2)},s} + \delta^{p_1^{(3)},com} \cdot W \end{cases} \quad (9)$$

The full derivative is the sum of the derivatives at each of the nodes, $\partial E / \partial W = \sum_{i=1}^{n} \partial E / \partial W_i$. The procedure for back-propagation and gradient descent are presented in Algorithm 2.

TABLE 3
BACK-PROPAGATION AND GRADIENT DESCENT PROCEDURE IN RNKN

**Algorithm 2**: Back-propagation mechanism and gradient descent of RNKN
**Input:** $Tree$: a Huffman knowledge tree
**Output:** $List_{error}$: an error list, including softmax errors, downward errors, and complete errors
   $Weight$: updated weight matrix
   $Weight_s$: updated softmax weight matrix
   $List_{vector}$: updated list of node vectors
***Begin***
1: **Initialize** the step of vector $Step_{Vector}$, weight matrix $Step_{Weight}$, and disease classification matrix $Step_{Weight_s}$, respectively.
2: ***Function*** BackPropagation($node$):
3:   $y_{node} = Weight_s \cdot Vector_{node}$
4:   $Error_{node,softmax} = (y_{node} - t_{node}) \cdot Weight_s \cdot f'(Vector_{node})$
5:   **if** $node == root$ **then** : $Error_{node,complete} = Error_{node,softmax}$
6:   **else**: $Error_{node,complete} = Error_{node,softmax} + Error_{parent,down}$
7:   $Error_{node,down} = Weight \cdot Error_{node,complete} \cdot f'(Vector_{node})$

```
 8:    BackPropagation(node.getLeftChildren())
 9:    BackPropagation(node.getRightChildren())
10: end Function
11: Function Gradient(Tree):
12:    for node_i ∈ Tree do
13:       child_{left} = node_i.getLeftChildren()
14:       child_{right} = node_i.getRightChildren()
15:       Vector_{combine}[v_1,...,v_d,v_{d+1},...,v_{2d}] ← [child_{left}.getVector(), child_{right}.getVector()]
16:       J_{weight} += Error_{node_i,complete} · Vector_{combine}
17:       J_{weights_s} += (y_{node_i} − t_{node_i}) · f(Vector_{node_i})
18:       if node_i ∈ leaf node then
19:          Vector_{node_i} −= Step_{vector} · Error_{node_i,softmax}
20:          Vector_{node_i} ← Vector_{node_i}/‖Vector_{node_i}‖
21:       end if
22:    end for
23:    Weight −= Step_{weight} · J_{weight}
24:    Weight_s −= Step_{weights_s} · J_{weights_s}
25: end Function
```

The forward phase starts from leaf nodes, which are initialized to a series of random vectors. The backward phase follows the topology of the Huffman knowledge tree beginning at the root-node. For each tree in the training set, one recursive epoch requires exactly one forward and one backward phase. Until the full loss of cross entropy less than a threshold or the number of epochs more than a specified count, RNKN converges the weight matrix $W$, the softmax weight matrix $W_s$, and knowledge embedding $Vector$ to an optimum state.

## 4 EXPERIMENTS AND RESULTS

### 4.1 Diagnosis task and corpus

In this study, we conducted multiple experiments to verify the effectiveness of RNKN on an actual CEMR dataset, with the aim of diagnosing disease according to symptoms and historical diseases of a patient. Compared with most of the existing diagnostic models, which are applied to single-disease in a restricted domain, we present the performance of RNKN for whole departments.

As described in Section 2.1 above, we chose the manually annotated 2988 CEMRs with the help of medical professionals and kept only the discharge summaries and progress notes as the sources of knowledge. Referencing the medical concept annotation guideline and the assertion annotation guideline given by i2b2 [40], we extracted the disease entity and the symptom entity from five kinds of annotated entities as evidence, and selected symptom–disease relation as knowledge. In addition, we adopted the modifiers *present*, *occasional*, *conditional*, *historical*, *possible*, *absent*, and *not associated with the patient*, to represent the condition of entity. Following completion of the above steps, the whole CEMRs were randomly divided into 2362 training samples and 626 testing samples. Furthermore, 2656 entities that consist of 1690 disease entities and 1742 symptom entities are extracted from training corpus. Finally, we employ 8682 pieces of knowledge to infer the possibility of 371 diagnosable diseases.

### 4.2 Quantitative evaluation of the training process

In this section, we discuss selection of the optimal trained RNKNs for multi-disease diagnosis. To describe the diagnosis result and the fitting degree of trained models, we adopt $\bar{P}@10$, discounted cumulative gain (DCG) [41], and loss function for quantitative evaluation. When the actual disease contained in the first ten diagnosis results is considered as a correctly diagnosed case, $\bar{P}@10$ is defined as the ratio of correctly diagnosed cases to the total number of cases: $\bar{P}@10 = r/n$. The DCG is a classical measurement in information retrieval, of which score can be calculated by $DCG = rel_1 + \sum_{i=2}^{n} rel_i/\log_2 i$. $rel_i$ represents the relevance between the $i$th candidate disease and the actual disease; a correct diagnosis is "1," whereas a misdiagnosis is "0." $\bar{P}@10$ measures whether the actual disease can be obtained in the first ten results. In contrast to $\bar{P}@10$, we use DCG to measure whether the actual disease can be ranked on the top of candidate lists as much as possible. The higher the ranking of an actual disease, the greater DCG score can be calculated. Moreover, we employ the loss function of cross entropy as the last indicator, which evaluates directly the fitting degree of the trained model. Detailed results of the training process are provided in Fig. 3.

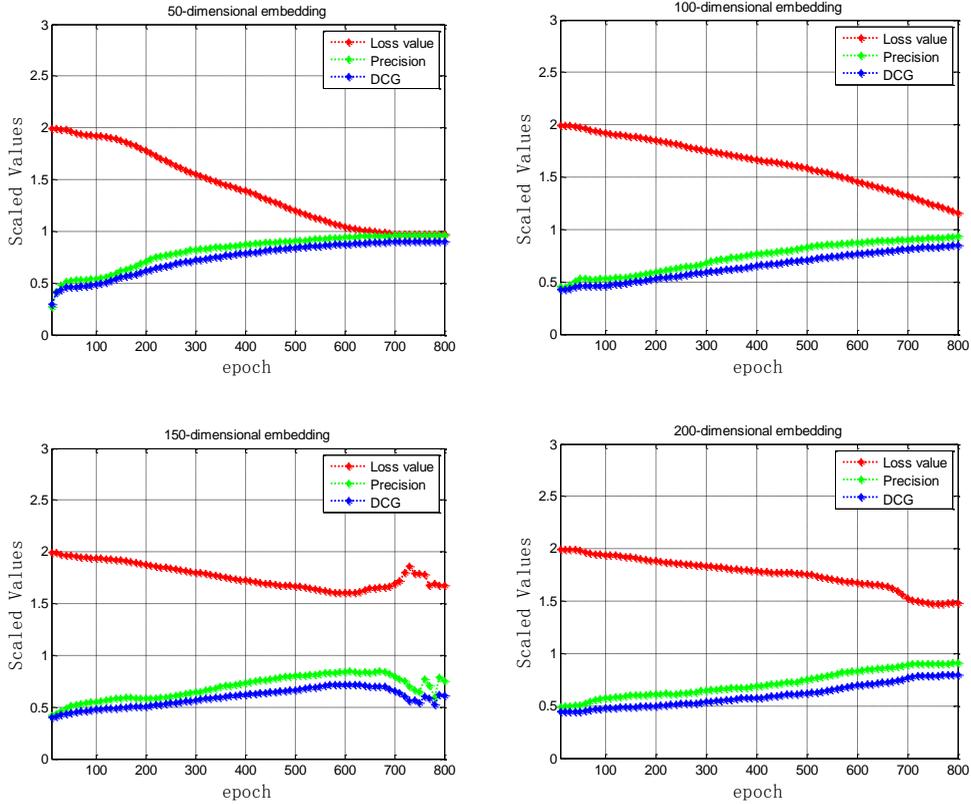

Fig. 3. Training results of RNKN with four different dimensional embeddings.

In the experiments conducted, we defined different dimensionalities for the basic vector: [50, 100, 150, 200]. All vectors and weight matrixes were initialized by randomly sampling each value from a uniform distribution: $U(-r, r)$, where $r = 0.1$. The gradient steps were initialized empirically to 0.001. After 800 computation iterations, the curves of precision and DCG both gradually tended to 1.0. Conversely, the values of the loss function decreased as the training of RNKN deepened, and finally inclined to a relatively stable state. As shown in Fig. 3, we selected the trained models generated in the 800th epoch as the final candidates for the 50, 100, and 200 dimensional RNKN. Because of the fluctuation of the 150-dimensional RNKN in the end stage, we synthesized three indicators to determine the peak point (670th epoch) as the best candidate model for diagnostic tasks.

### 4.3 Effectiveness of RNKN in multi-disease diagnosis

To verify the effectiveness of RNKN in multi-disease diagnosis, eight models were compared as follows.
- Five machine learning models: Naive Bayes, Logistic regression, Neural Network, J48, and ZeroR were considered as baselines. Because of the incapability of some models to conduct multi-classification, we built a two-class processing for each disease. Consequently, we ranked the candidate diseases by confidence and evaluated the validity of the models.
- Markov logic network (MLN): MLN [42] is a unified inference framework that combines first-order logic knowledge and probabilistic graphical models. This was used to compare the performance of RNKN with a common knowledge base. We employed Tuffy [43] as the base solver.
- RNKN: We used the methodology of RNKN, as mentioned in Section 3. The whole symptom and disease entities were extracted as evidence from CEMR, whatever the entity modifier is.
- RNKN+: we used the same setup as basic RNKN, but only symptom and disease entities with "*present*" modifier were considered as evidences.

Owing to the incompleteness of annotated entities, some CEMRs have no available evidence, because of the limitations of the modifier. Apart from this exception, there were 528 available CEMRs out of 626 testing samples. Fig. 4 shows the performance of six baselines and RNKN on 528 CEMRs. In the experiments, we found that machine learning methods are difficult to apply for multi-disease diagnosis, especially when a patient's evidence is sparse. In addition, the experimental results showed that MLN outperformed the majority of machine learning approaches. However, there is still a certain gap between the overall effect and expected results of MLN. Analysis of the detailed output results shows that a possible cause is MLN being unable to provide enough candidate diseases when CEMR has a deficiency of knowledge in regard to the given evidence. If the number of candidates is less than 10, the diagnostic accuracy would inevitably be reduced in three indicators. Synthesizing the results of four RNKNs, we found that the larger the dimension of embedding is, the worse is the performance of RNKN. It can be explained intuitively by the problem of parameter selection. When the number of parameters rises rapidly with the increasing dimension, the optimization process of RNKN becomes harder and more costly by adjusting network parameters. Nonetheless, the effectiveness of RNKN is obviously superior to that of MLN using the same medical knowledge base. $\bar{P}@5$, $\bar{P}@10$, and DCG of RNKN@50 exceeded 0.55, 0.67, and 0.59, respectively.

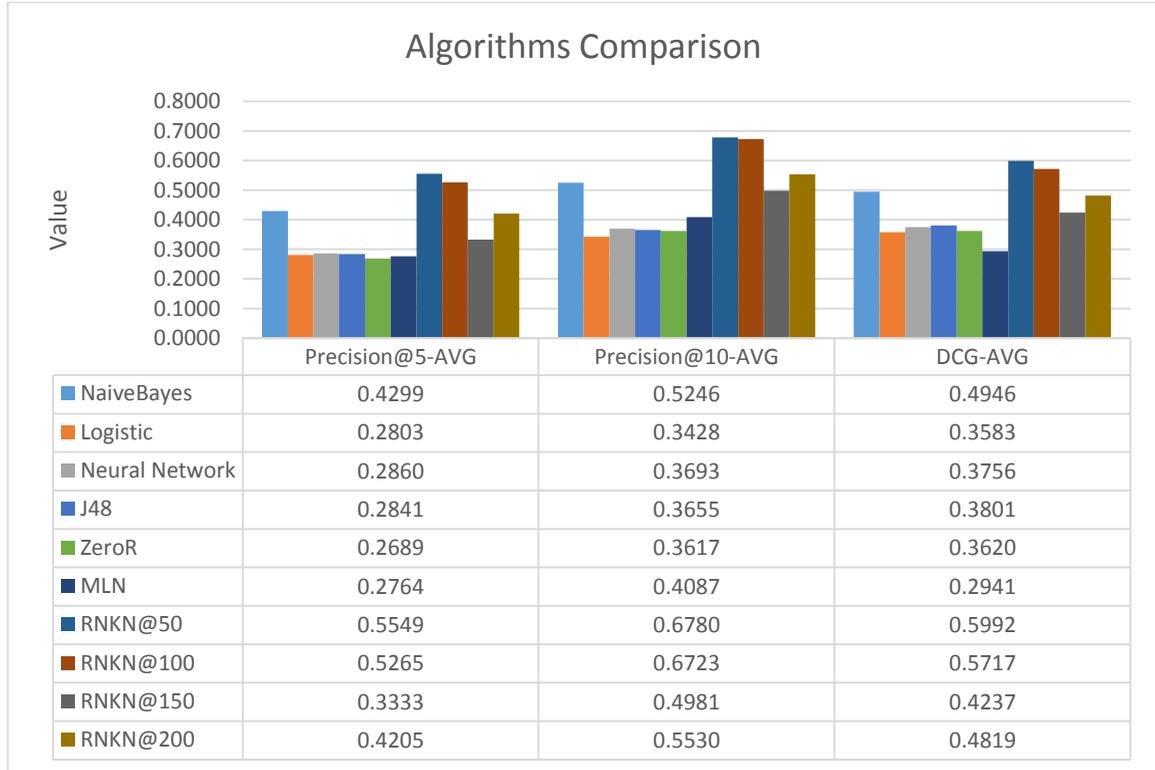

Fig. 4. Comparison of seven diagnostic algorithms with looser evidence: five machine learning methods, Markov logic network (MLN), and recursive neural knowledge network (RNKN).

In contrast to basic RNKN, the evidence of RNKN+ is restricted to "*present*" entities. Because "*present*" entities can emphasize the "current" physical condition of a patient, the extraction strategy is consistent with the commonsense of medical practice. In Fig. 5, it can be seen that RNKN+ consistently outperforms other baselines. Compared to the same tests with the basic RNKN@50, RNKN+@50 has significantly improved by 6% $\bar{P}@5$, 4% $\bar{P}@10$, and 4% DCG. This illustrates that the explicit evidence is more beneficial to make inferences based on knowledge, while also demonstrating the effectiveness of RNKN in multi-disease diagnosis.

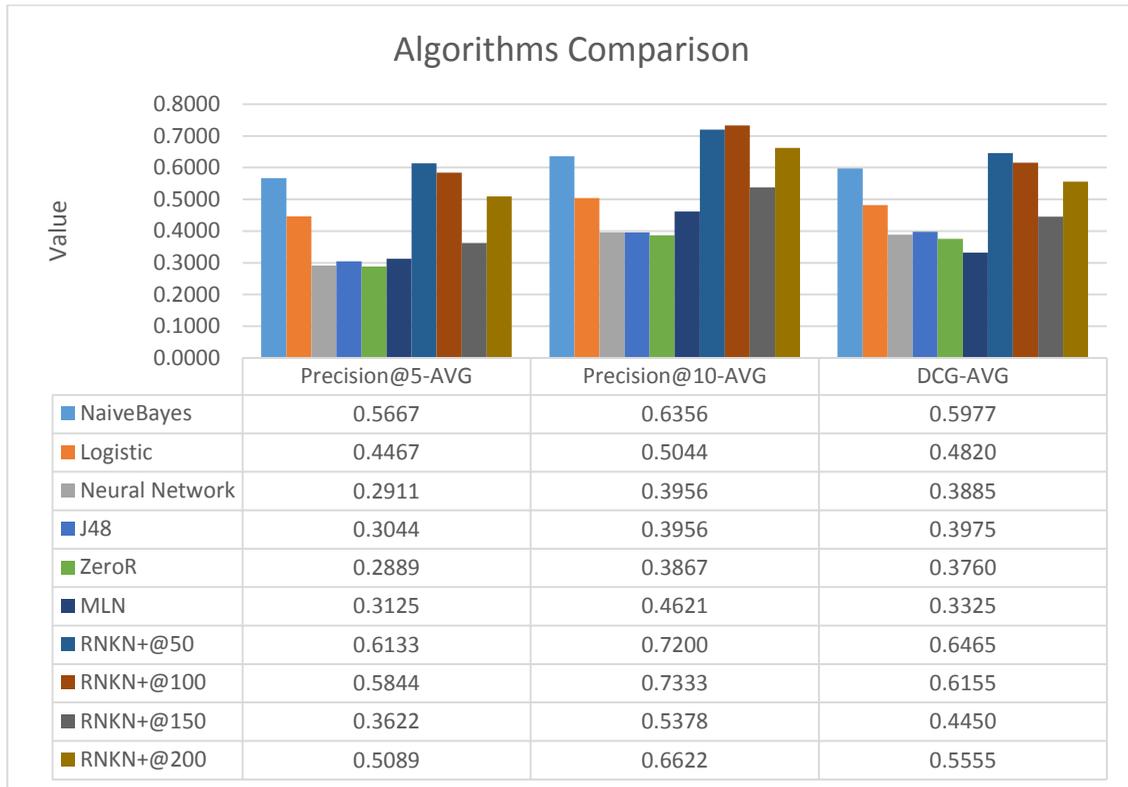

Fig. 5. Comparison of seven diagnostic algorithms with strict evidence: five machine learning methods, Markov logic network (MLN), and recursive neural knowledge network (RNKN+).

| | Precision@5-AVG | Precision@10-AVG | DCG-AVG |
|---|---|---|---|
| NaiveBayes | 0.5667 | 0.6356 | 0.5977 |
| Logistic | 0.4467 | 0.5044 | 0.4820 |
| Neural Network | 0.2911 | 0.3956 | 0.3885 |
| J48 | 0.3044 | 0.3956 | 0.3975 |
| ZeroR | 0.2889 | 0.3867 | 0.3760 |
| MLN | 0.3125 | 0.4621 | 0.3325 |
| RNKN+@50 | 0.6133 | 0.7200 | 0.6465 |
| RNKN+@100 | 0.5844 | 0.7333 | 0.6155 |
| RNKN+@150 | 0.3622 | 0.5378 | 0.4450 |
| RNKN+@200 | 0.5089 | 0.6622 | 0.5555 |

To further verify the diagnostic effects of RNKN, the detailed DCG scores (y-axis) plotted against the serial numbers of 528 CEMRs (x-axis) are presented for six comparison models and RNKN+ in Fig. 6 and Fig. 7. When the actual diagnosed disease is ranked at the first item of the candidate list, the DCG score is "1." Conversely, when the candidate list is incomplete and does not contain the actual diagnosed disease (such as MLN), the DCG score is "0."

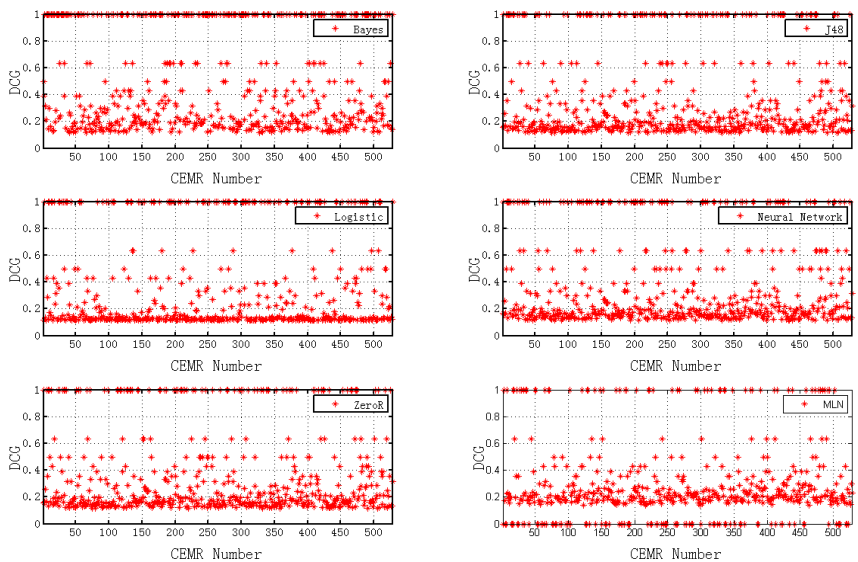

Fig. 6. Discounted cumulative gain (DCG) of machine learning models and Markov logic network (MLN).

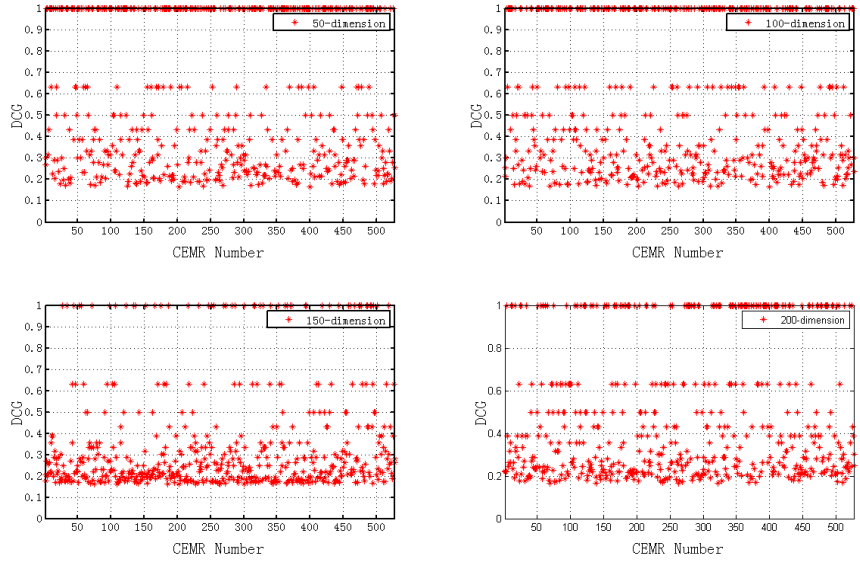

Fig. 7. Discounted cumulative gain (DCG) of RNKN+ with four different dimensional embeddings.

### 4.4 Qualitative evaluation of interpretable representations

To qualitatively assess the interpretability of the learned representations, we plotted 200-dimensional embeddings of 1690 disease entities in a 2D space using t-SNE [44]. As shown in Figs. 8-11, the scatterplot of these entity embeddings presents cluster characteristics. A scatterplot of entity embedding with community structures indicates that the entities within a common group have a highly semantic relevance. To depict the evolutionary process, we present a beginning fragment (10th epoch) and an ending fragment (800th epoch) in the training period. Along with the increasing number of iterations, the community border will become gradually clear.

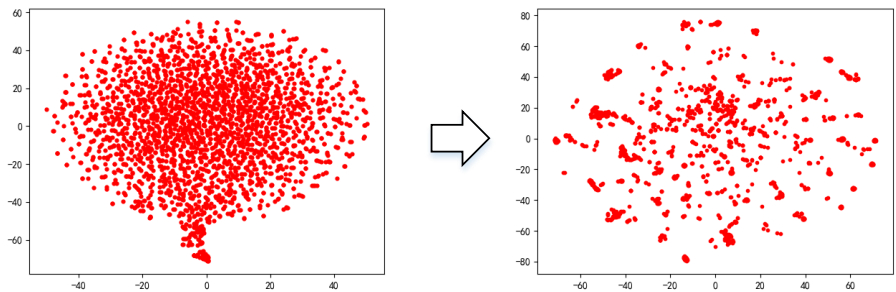

Fig. 8. Evolutionary scatterplot of 50-dimensional RNKN+

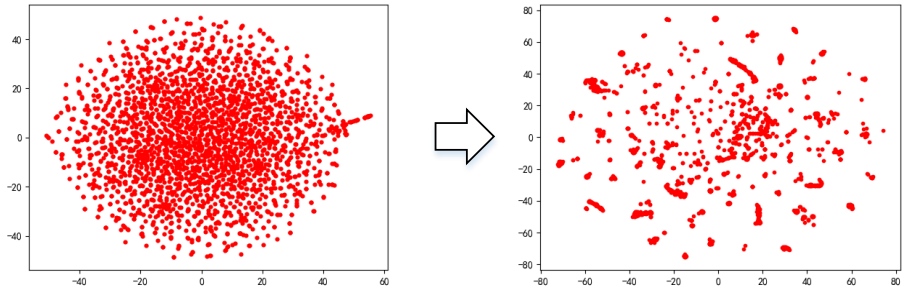

Fig. 9. Evolutionary scatterplot of 100-dimensional RNKN+

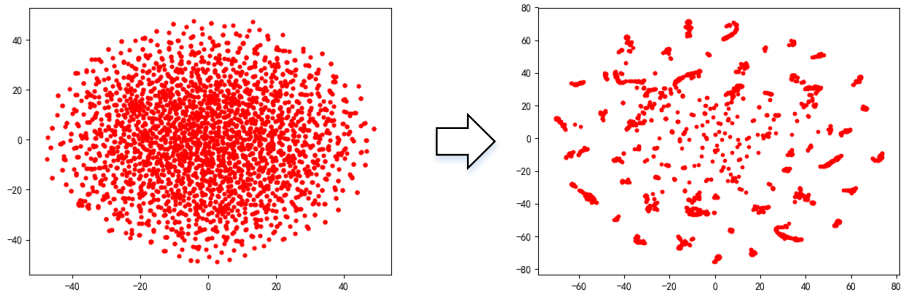

Fig. 10. Evolutionary scatterplot of 150-dimensional RNKN+

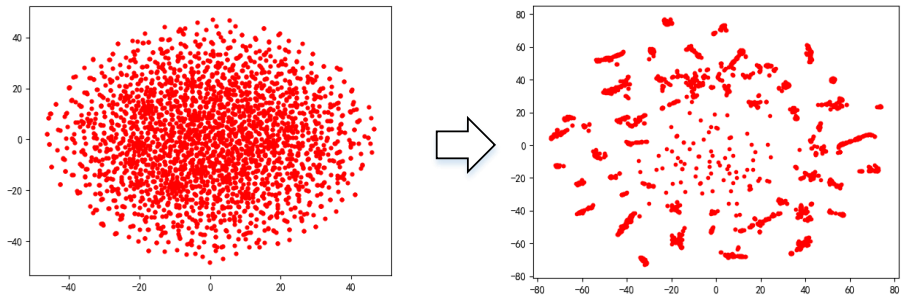

Fig. 11. Evolutionary scatterplot of 200-dimensional RNKN+

To explain the actual meaning of each cluster, we mapped each disease embedding of 200-dimensional RNKN+ into an International Classification of Diseases (ICD) code. According to ICD10 code hierarchy, the ICD10 codes of entities with the same first two characters are considered as the same kind of disease and are classified into the same category. In Fig. 12, the colors of the dots represent the disease categories and the text annotations describe the detailed categories in the ICD10 hierarchy. Despite the appearance of exception nodes in some clusters, most of the nodes can be grouped into the appropriate categories. It is clear that RNKN exhibits interpretable embedding that is well aligned with the semantic information of medical terms.

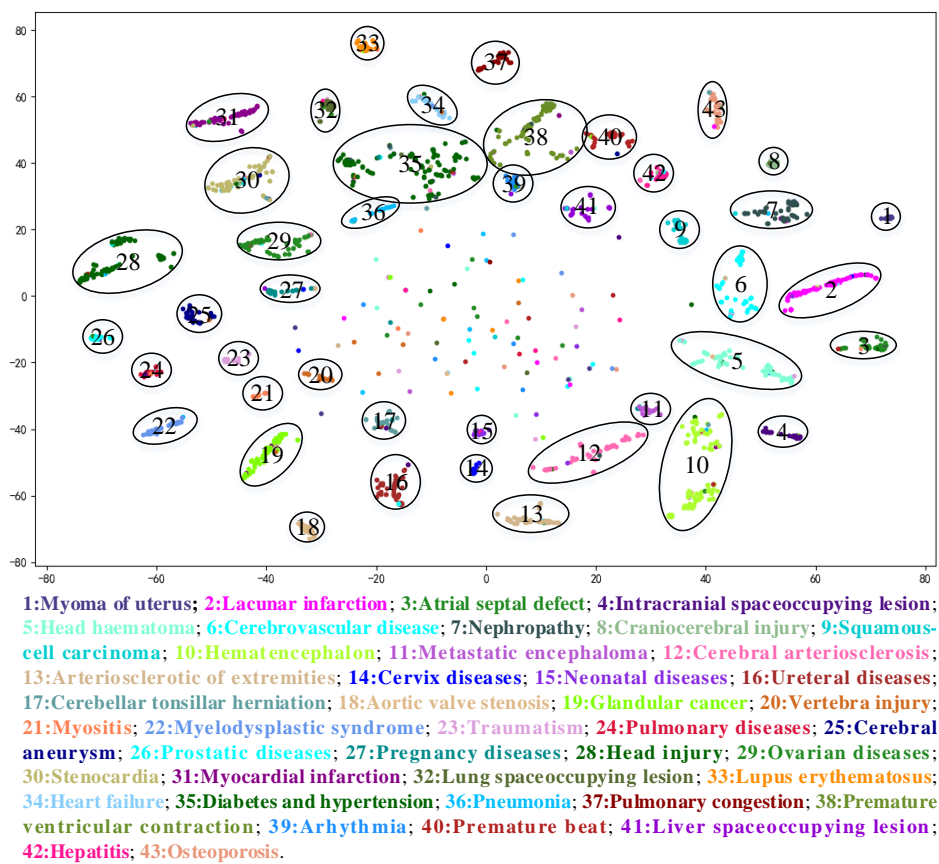

**1:**Myoma of uterus; **2:**Lacunar infarction; **3:**Atrial septal defect; **4:**Intracranial spaceoccupying lesion; **5:**Head haematoma; **6:**Cerebrovascular disease; **7:**Nephropathy; **8:**Craniocerebral injury; **9:**Squamous-cell carcinoma; **10:**Hematencephalon; **11:**Metastatic encephaloma; **12:**Cerebral arteriosclerosis; **13:**Arteriosclerotic of extremities; **14:**Cervix diseases; **15:**Neonatal diseases; **16:**Ureteral diseases; **17:**Cerebellar tonsillar herniation; **18:**Aortic valve stenosis; **19:**Glandular cancer; **20:**Vertebra injury; **21:**Myositis; **22:**Myelodysplastic syndrome; **23:**Traumatism; **24:**Pulmonary diseases; **25:**Cerebral aneurysm; **26:**Prostatic diseases; **27:**Pregnancy diseases; **28:**Head injury; **29:**Ovarian diseases; **30:**Stenocardia; **31:**Myocardial infarction; **32:**Lung spaceoccupying lesion; **33:**Lupus erythematosus; **34:**Heart failure; **35:**Diabetes and hypertension; **36:**Pneumonia; **37:**Pulmonary congestion; **38:**Premature ventricular contraction; **39:**Arhythmia; **40:**Premature beat; **41:**Liver spaceoccupying lesion; **42:**Hepatitis; **43:**Osteoporosis.

Fig. 12. Scatterplot of the final representations of 200-dimensional RNKN+

## 5 CONCLUSIONS

In order to implement a knowledge-based neural network framework, we proposed and presented the recursive neural knowledge network (RNKN). In RNKN, the mathematical derivation of back-propagation and forward-propagation is rigorously deduced for calculating cross-entropy error and knowledge embedding. To address the problem of multi-disease diagnosis, we manually annotated 8682 pieces of medical knowledge from Chinese Electronic Medical Records (CEMRs). Then, a Huffman tree based on first-order logic knowledge was constructed with both symptom and historical disease entities as inputs, and the normalized distribution of diagnosable diseases as output. After 800 epochs on a real CEMR dataset, four different dimensional RNKNs were trained until they met the overall convergence criteria. Through comparison with other algorithms, the effectiveness and promise of RNKN in multi-disease diagnosis was demonstrated. Empirical tests with actual records illustrate that RNKN can improve diagnostic accuracy, especially when more explicit evidence is extracted. Mapping trained high-dimensional vectors into a 2D space, we further confirmed the interpretability of entity embedding.

RNKN is a knowledge-based neural network framework that is applicable to many AI problems for future development. Future work will involve intelligent healthcare, in which we plan to fuse the representation ability of neural network models and the inference ability of probabilistic graphical models with huge amounts of medical knowledge, and investigate the application of RNKN in a variety of domains.

## ACKNOWLEDGMENT

The Chinese Electronic Medical Records used in this paper were provided by Second Affiliated Hospital of Harbin Medical University.

**Funding:** This work was supported by the National Natural Science Foundation of China [grant numbers 71531007, 71622002, and 71471048].